\documentclass{article}
\usepackage{spconf,amsmath,epsfig}
\usepackage{footnote,lipsum,hyperref}

\begin{document}\sloppy

% Example definitions.
% --------------------
\def\x{{\mathbf x}}
\def\L{{\cal L}}

% Title.
% ------
\title{Incremental Learning Algorithm for Sound Event Detection}
%
% Single address.
% ---------------
% \name{Anonymous ICME submission}
%Address and e-mail should NOT be added in the submission paper. They should be present only in the camera ready paper. 
\twoauthors
 {Eunjeong Koh\sthanks{Work performed as an intern at Qualcomm Technologies}}
	{University of California, San Diego\\
	Department of Music\\
	La Jolla, CA, USA\\
	\texttt{eko@ucsd.edu}}
 {Fatemeh Saki, Yinyi Guo, Cheng-Yu Hung, Erik Visser}
	{Qualcomm Technologies, Inc.,\\
	San Diego, CA, USA\\
	\texttt{\{fsaki,yinyig,chengyuh,evisser\}}\\
	\texttt{@qti.qualcomm.com}}

\vspace{3mm}
\address{}

\maketitle

\begin{abstract}
This paper presents a new learning strategy for the Sound Event Detection (SED) system to tackle the issues of i) knowledge migration from a pre-trained model to a new target model and ii) learning new sound events without forgetting the previously learned ones without re-training from scratch. In order to migrate the previously learned knowledge from the source model to the target one, a neural adapter is employed on the top of the source model. The source model and the target model are merged via this neural adapter layer. The neural adapter layer facilitates the target model to learn new sound events with minimal training data and maintaining the performance of the previously learned sound events similar to the source model. Our extensive analysis on the \textbf{DCASE16} and \textbf{US-SED} dataset reveals the effectiveness of the proposed method in transferring knowledge between source and target models without introducing any performance degradation on the previously learned sound events while obtaining a competitive detection performance on the newly learned sound events.

\end{abstract}

\section{Introduction}
\label{sec:intro}

Sound Event Detection (SED) is a rapidly growing research area that aims to analyze and recognize a variety of sound events in a continuous audio signal. Neural Networks based methods such as Convolutional Neural Networks (CNNs) have recently been used for SED systems to advance the performance of these systems \cite{mesaros2018multi,mesaros2016tut}. In the Detection and Classification of Acoustic Scenes and Events (DCASE) Task 4 \cite{DCASE2019Task4}, the state-of-the-art SED systems have been tested using real data that is either weakly labeled or unlabeled and simulated strongly labeled data with the onset and offset times of sound events. 

\vspace{2mm}

Although SED problem has been attracting many researchers, a vast majority of the state-of-the-art systems are focused on advancing the performance of the SED systems by utilizing weakly labeled data \cite{DCASE2019Task4}. To our knowledge, one of the important and non-investigated challenges of the current SED models is their closed-set nature, where a fixed and limited number of known classes are used during the training. It is difficult to collect exhaustive training samples or to properly annotate all the training data to train the classifiers. Hence in the closed-set classifiers, only a limited number of classes are considered for training, with the assumption that during test time, the test data is drawn from the same set of classes as the training data. However, the SED systems in nature are open-set problems, in other words, the test data could include samples associated
with unknown sound events as well. Therefore, it is always desired to have a flexible model that can learn new classes, once new training data including new sound events becomes available. Then again, it is required to still remember the previously learned classes after adopting the new classes, and learning capability is referred to as continuous learning or incremental learning. 

One of the main challenges associated with these types of continuous learning algorithms is catastrophic forgetting \cite{mccloskey1989catastrophic}. That is if the information about the previously learned categories is unavailable when a new task is added, it overwrites the previously learned information. Hence, it leads to the performance degradation of past tasks. The ability of continuous/incremental learning over time represents a long-standing challenge for Machine Learning and Neural Networks \cite{thrun1995lifelong}. Recently, in the areas of computer vision and natural language processing \cite{goodfellow2016deep,chen2019transfer}, Transfer Learning (TL) has shown great potential to 1) identify the transferable knowledge by accommodating new knowledge and 2) retain previously learned information. Some recent works have explored TL for audio applications \cite{cramer2019look,kumar2018knowledge,jung2019polyphonic,saki2019open}, which focus on knowledge transfer between databases with various qualities, mismatch downstream tasks, and domains. However, it remains to be seen how a flexible TL model for SED task to audio knowledge transfer can be done.

%However, it is still desired to have a flexible model for SED tasks that can transfer the previously learned knowledge effectively while maintaining the performance of the pre-learned event classes.

% For designing a flexible model, recent study in natural language processing proposed a neural model for sequence labeling to transfer knowledge from source models using additional neurons in the output layers \cite{chen2019transfer}. This neural adapter can transfer the previously learned knowledge effectively while maintaining the performance of the pre-learned event classes. However, it remains to be seen how a direct adaptation approach for SED task to audio knowledge transfer can be done.

\begin{figure*}[h]
\centering
\includegraphics[width=10cm, height=8cm]{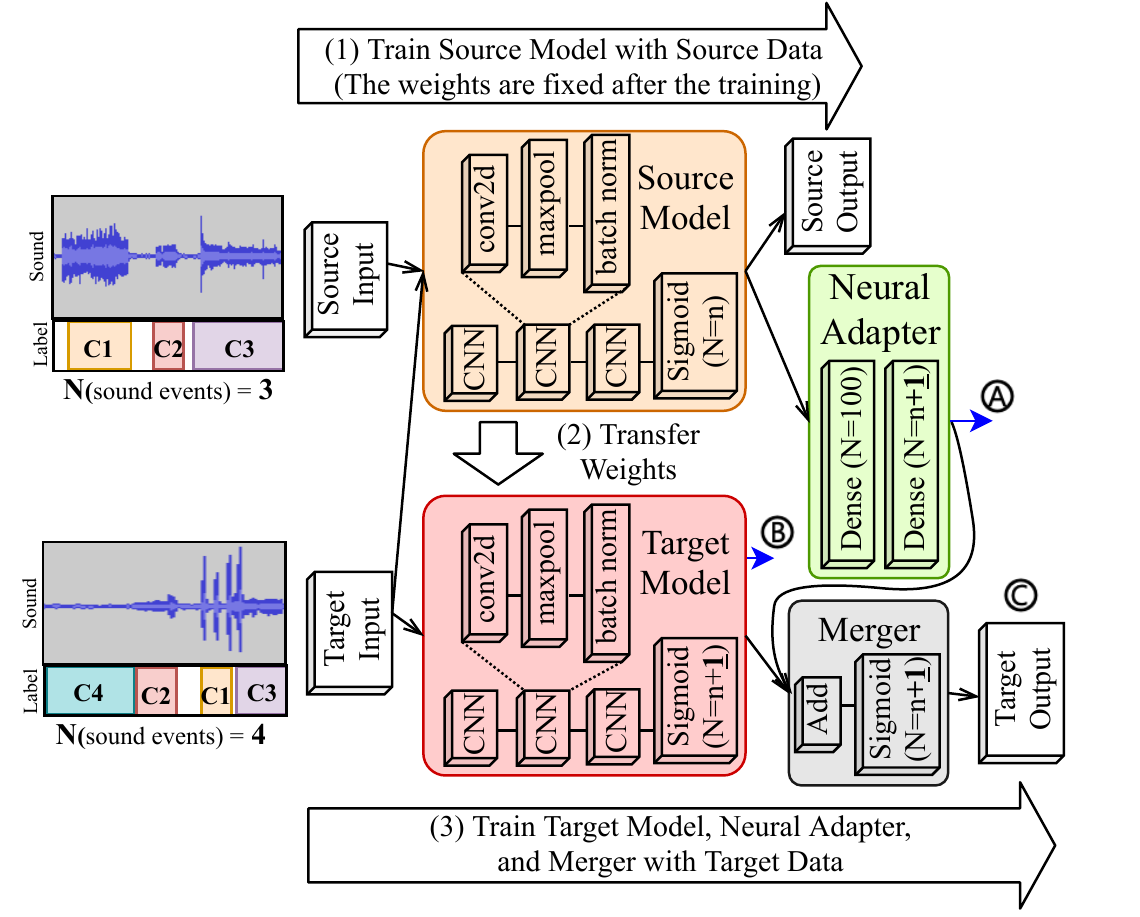}
  \caption{\textbf{SED incremental learning algorithm structure.} Source input consists of \textit{N} sound events and target input consists of \textit{N+1} sound events leaving one sound event out to be incrementally learned later with the target model. The source and target model includes three 2D convolution layers, 2D max-pooling operation, and batch normalization layers. We use a sigmoid function for output activation of the source and the target model. Source model's weights are fixed during target model training. The target model is initialized with the optimal weights of the source model. The neural adapter consists of fully-connected dense layers. Note that source and target models have different output dimensions due to the new class in the target data. \textcircled{A}\& \textcircled{B} are intermediate outputs for analysis, we add \textcircled{A} and \textcircled{B} to \textcircled{C} using merger, and \textcircled{C} is the final output of our model.}
  
  \vspace{+3mm}
  \label{fig:adapter_structure}
\end{figure*}

% \vspace{3mm}

In this paper, we present an incremental learning algorithm for SED applications effectively transferring knowledge from a source model to a target model. Our method updates the target model when new sound events are available without any catastrophic forgetting. Motivated by the recent TL advances in natural language processing \cite{chen2019transfer}, we utilize a neural adapter to bridge the gap between the source model and the target model. We combine new neurons for transferring parameters from the source model and implement a neural adapter to lessen the gap between the source and the target data distribution. For testing the performance of the neural adapter, we test the basic simple TL approach, then we show the impact of our proposed neural adapter for the SED task, while testing several TL options on just one category. The results show that our method provides an effective knowledge transfer mechanism between source and target domains without any additional training examples and any performance degradation of the previously learned tasks in the source domain. Our learning algorithm helps to transmit the predictions from the source model into that of the target model.

\vspace{2mm}

The rest of the paper is organized as follows. In section 2, we cover the proposed SED incremental learning algorithm, followed by the experimental results in section 3. The conclusion and the discussion are provided in section 4.

\vfill

\section{SED Incremental Learning Algorithm}
\label{sec:sed_problem}

\subsection{SED Problem Formulation}

Given an input audio file that includes several acoustic scenes, a standard SED predicts the corresponding labels that indicate the annotation of all the sound events in the scenes. The time-series audio input is represented by audio embedding vectors and the event label file includes the information of event specification, such as the sound event onset/offset time and sound event label.

\subsection{Transfer Learning (TL) Workflow}

Our TL mechanism consists of several steps (see Figure \ref{fig:adapter_structure}); (i) a pre-trained source model, $M_S$, for a certain number of categories, in the source domain, $D_S$; (ii) a target model, $M_T$, that utilizes the source model $M_S$ parameters as a starting point to learn the target domain, $D_T$. In addition to the typical TL mechanism, which is (i) and (ii), we add a neural adapter (iii) a connection system between the source model $M_S$ and the target model $M_T$ for effectively transferring knowledge from the source domain $D_S$ to the target domain $D_T$. This connection will alleviate the effect of information discrepancy and prevent any catastrophic forgetting on the previously learned information. Finally, we (iv) jointly train the target model $M_T$ and the connection system together to effectively learn the target domain $D_T$ information. Note that the parameters of the source model $M_S$ are not updated during this target model $M_T$ training process.

\subsection{Model Architecture}
\label{sec:method}

\subsubsection{SED Source Model}

Figure \ref{fig:adapter_structure} describes the process and the structure of our SED incremental learning methodology. We revise the Convolutional Neural Network (CNN) proposed by Salamon and Bello \cite{salamon2017deep}, which includes three 2D convolution layers, a 2D max-pooling operation, and batch normalization layers. Each layer processes 64 convolutional filters. The input to the network is a Mel spectrogram of size 128x128 that is extracted from a one-second audio file. ReLU activation functions are applied to the convolutional layers %, $f(x) = max(0,x)$, 
to reduce the backpropagation errors and accelerate the learning process \cite{goodfellow2016deep}. Sigmoid function %, $f(x) = \frac{1}{1+e^{-x}}$
are used as the output activation function with \textit{N} classes. Adam optimizer \cite{kingma2014adam} and binary cross-entropy loss function are used. The stopping criterion is set as 500 epochs with an early-stopping rule, if there is no improvement to the F1 score during last 100 learning epochs \cite{mesaros2016metrics}. The final model has 720k parameters. This model is implemented in Keras \cite{chollet2015keras}.

\subsubsection{Incremental Learning using Neural Adapter}
\label{sec:adapter}

In this work, the $M_S$ is trained for \textit{N} sound events using the aforementioned CNN model. The goal is to create a $M_T$ for \textit{N+1} sound events without training from scratch. Note that the \textit{N} sound events are common for both $M_S$ and $M_T$. To this end, the $M_T$ has the same CNN structure as the $M_S$ and the trained parameters of the previously learned \textit{N} sound events are utilized from the $M_S$ as an initial training point for the $M_T$. To account for the new category in the $M_T$, we modify its output layer with \textit{N+1} sigmoid activation. It is well known that learning a new task via such a simple transfer learning paradigm usually results in forgetting the previously learned classes while adding new classes. To avoid this knowledge-lost problem, we adopt the TL mechanism proposed by Chen and Moschitti \cite{chen2019transfer}. In this method, to effectively transfer the knowledge between the $M_S$ and the $M_T$, a neural adapter is utilized to bridge the two models and jointly trained with the $M_T$. More specifically, a neural adapter consists of two fully-connected dense layers over the last layer of $M_S$ is used to connect the $M_S$ to the $M_T$. This process is called the element-wise summation which integrates the outputs from the source and target domain and finally processes \textit{N+1} categories. The parameters of the neural adapter and the $M_T$ are learned simultaneously while the $M_S$ parameters are fixed.

\section{Experimental Results}
\label{sec:experiments}

\subsection{Datasets}

We start our implementation with building $M_S$ using only \textit{N} sound events for $D_S$, out of the \textit{N+1} events, leaving one sound event out to be learned incrementally later with the $M_T$. We train the $M_S$ using the $D_S$ until the optimal parameters are achieved. These parameters and results will be saved and re-utilized for $D_T$. We evaluate our algorithm over three datasets; the DCASE 2016 challenge Task 2 (\textbf{DCASE16}) \cite{mesaros2016tut}, the UrbanSound-SED (\textbf{US-SED}), and UrbanSound-8K (\textbf{US-8K}) \cite{salamon2014dataset} dataset.  

\vspace{2mm}

The \textbf{DCASE16} dataset includes eleven different sound events for the SED challenges\footnote{\url{http://www.cs.tut.fi/sgn/arg/dcase2016/task-sound-event-detection-in-synthetic-audio}}, \textit{clearing throat, coughing, door knock, door slam, drawer, human laughter, keyboard, keys (put on the table), page turning, phone ringing, and speech}. In this work, we use four sound events; \textit{``door knock", ``door slam", ``keyboard",} and \textit{``phone ringing"}. To generate soundscapes from these sound files, we use the Scaper open-source library \cite{salamon2017scaper} for the synthesis and augmentation \footnote{We revise this open-source implementation: \\
\url{https://github.com/justinsalamon/scaper\_waspaa2017}}.  We create 800 soundscapes for $M_S$ training data and 200 soundscapes for each test/validation data. The duration of each soundscape audio file is ten seconds. It is worth noting that the soundscapes are generated in such that each event appears at least once and a maximum of two times in every soundscape file. This \textbf{DCASE16} dataset denotes a clean and well-labeled dataset for our experimental setting.

\vspace{3mm}

The \textbf{US-SED} dataset \cite{salamon2014dataset} includes ten different sound classes; \textit{air conditioner, car horn, children playing, dog bark, drilling, engine idling, gun shot, jackhammer, siren,} and \textit{street music}. In this work, we use the pre-generated UrbanSound soundscape audio files from the Scaper study \cite{salamon2017scaper}. It has 10,000 soundscape files with a duration of ten seconds. Each of the soundscape files has a minimum number of sound events as zero, and the maximum number of nine. Thus, some of the soundscapes might be empty of the sound events of interest. This situation can be interpreted as a more realistic SED framework compared to the aforementioned \textbf{DCASE16}, where each of its soundscape files contains all the sound events. For our experiment, we considered five sound events; \textit{``car horn", ``dog bark", ``gun shot", ``siren",} and \textit{``street music"}. 4,995 files are used for $M_S$ training, and 1,665 files for each of the test and verification data. 

\begin{table}[]

\caption{\textbf{F1-score of the $M_S$, Simple TL, and Neural Adapter TL over different settings on DCASE16.} Simple TL means $M_T$ built based on a simple (or typical) TL method without the neural adapter, and Neural Adapter TL means $M_T$ built using the neural adapter TL method in addition to the simple TL. The $M_S$ sections shows the F1-score of $M_S$ for different $D_S$ domain settings. Simple TL section illustrates the F1-score of the $M_T$ that are built via a simple TL approach. Neural Adapter TL section reports the F1-score of the proposed approach. $D_S$ columns of the Simple TL and Neural Adapter TL indicate the F1-score of $M_T$ on the previously learned classes. The New columns present the F1-score of the newly learned class and All columns report the overall F1. Rows in the table depict each of the test scenarios assuming that unseen label in $D_S$ is newly introduced in $D_T$. DCASE16 includes four sound event classes, C1: keyboard, C2: door slam, C3: phone ringing, and C4: door knock.}

% \footnotesize
\centering
\setlength\tabcolsep{1.5pt}

\vspace{3mm}

\begin{tabular}{|c|c|c|c|c|c|c|c|}
\hline
\textbf{DCASE16}         & $M_S$         & \multicolumn{3}{c|}{Simple TL}                   & \multicolumn{3}{c|}{Neural Adapter TL}           \\ \hline
% \multirow{2}{*}{Labels in $D_S$} & \multirow{2}{*}{\scriptsize{$D_S$}}    & \multicolumn{3}{c|}{\scriptsize{$D_T$ ($C_1$ $C_2$ $C_3$ $C_4$)} }   & \multicolumn{3}{c|}{\scriptsize{$D_T$ ($C_1$ $C_2$ $C_3$ $C_4$)}}    \\ \cline{3-8} 
 {Labels in $D_S$}               &  $D_S$       & $D_S$          & New            & All            & $D_S$          & New            & All            \\ \hline
$C_1$ $C_2$ $C_3$            & .9444          & .8518          & .666           & .8055          & .944           & .6666          & .8518 \\ \hline
$C_1$ $C_2$ $C_4$            & .8888          & .7777          & .8888          & .8055          & .8886          & .8888          & .8888 \\ \hline
$C_1$ $C_3$ $C_4$            & .8518          & .8518          & 1.0            & .8888          & .8513          & .8             & .8388 \\ \hline
$C_2$ $C_3$ $C_4$           & .7407          & .8518          & 1.0            & .8888          & .74            & 1.0            & .8055 \\ \hline

Overall                  & .8561 & .8332 & .8888 & .8469 & .8559 & .8388 & .8462 \\ \hline

\end{tabular}
\vspace{2mm}

\label{tab:dcase16}
\end{table}

\subsection{Performance of the Incremental Learning}

We evaluate our algorithm in three different settings; i) evaluating the $M_S$ trained on $D_S$ with \textit{N} sound events, ii) evaluating the $M_T$ that is built via a simple TL on the $D_T$ with \textit{N+1} sound events, and iii) evaluating the model that is built utilizing the neural adapter, where the $M_T$ is merged with the $M_S$ through the neural adapter and trained on $D_T$. The results are reported in terms of F1-score (see Table \ref{tab:dcase16} and \ref{tab:us_sed_8k}). We perform experiments per every class assuming it is newly introduced in the $D_T$. Note that \textit{N} indicates three sound event classes for \textbf{DCASE16} dataset and \textit{N} indicates four sound event classes for the \textbf{US-SED} and \textbf{US-8K} dataset.

\vspace{2mm}

Table \ref{tab:dcase16} and Table \ref{tab:us_sed_8k} present the evaluation results for the \textbf{DCASE16} and \textbf{US-SED}, respectively. These tables include three sections; $M_S$, Simple TL and Neural Adapter TL. The $M_S$ section shows the F1-score of $M_S$ models for different $D_S$. Section Simple TL illustrates the F1-score of the $M_T$ that are built via a simple transfer learning approach. Section Neural Adapter TL reports the F1-score of the proposed approach. $D_S$ columns of the Simple TL and the Neural Adapter TL indicate the F1-score of the new models on the previously learned classes after the incremental learning process. The New columns present the F1-score of the newly learned class, and finally the All columns report the overall F1-score on the \textit{N+1} sound events. Rows in the table depict different test scenarios assuming that unseen label in $D_S$ is newly introduced in $D_T$ for incremental learning. 
\vspace{2mm}

$\bullet$ DCASE16: By paying attention to Table \ref{tab:dcase16} and the $D_S$ column in the $M_S$ section versus the $D_S$ column in the Simple TL section, it can be seen that learning sound classes via the simple TL approach, without the neural adapter, results in performance degradation in the previously learned sound events. This drop in performance indicates a catastrophic forgetting in the $D_T$ on the original three categories in $D_S$. On the other hand, the consistent F1-scores between the $D_S$ column in the $M_S$ section and the $D_S$ column in the Neural Adapter TL section is an illustration that the neural adapter can properly maintain the knowledge learned from $D_S$ while learning the new class in the $D_T$. This result proves that the neural adapter manages to mitigate the knowledge forgetting and enabling the model to update to the new domain. %

\begin{table}[]
\caption{\textbf{F1-score of the $M_S$, Simple TL, and Neural Adapter TL method over different settings on US-SED and US-8K.} This table is configured the same as Table \ref{tab:dcase16} with different datasets and sound classes. US-SED and US-8K include five sound event classes, C1: street music, C2: siren, C3: gun shot, C4: dog bark, C5: car horn).}

\vspace{3mm}

% \footnotesize
\centering
\setlength\tabcolsep{1.5pt}

\begin{tabular}{|c|c|c|c|c|c|c|c|} %|c|c|c|c|c|c|c|c|
\hline
\textbf{US-SED}         & $M_S$         & \multicolumn{3}{c|}{Simple TL}                    & \multicolumn{3}{c|}{Neural Adapter TL}           \\ \hline
% \multirow{2}{*}{Labels in $D_S$} &  \multirow{2}{*}{\scriptsize{$D_S$}}    & \multicolumn{3}{c|}{\scriptsize{$D_T$ ($C_1$ $C_2$ $C_3$ $C_4$ $C_5$)}} & \multicolumn{3}{c|}{\scriptsize{$D_T$ ($C_1$ $C_2$ $C_3$ $C_4$ $C_5$)}} \\ \cline{3-8} 
      Labels in $D_S$                   &   $D_S$        & $D_S$           & New            & All            & $D_S$          & New            & All            \\ \hline
$C_1$ $C_2$ $C_3$ $C_4$         & .5503          & .5782           & .6697          & .5965          & .5640          & .6962          & .5966 \\ \hline
$C_1$ $C_2$ $C_3$ $C_5$         & .5897          & .5918           & .4316          & .5598          & .589           & .5270          & .5789 \\ \hline
$C_1$ $C_2$ $C_4$ $C_5$         & .5985          & .6040           & .4493          & .5731          & .5947          & .4232          & .5666 \\ \hline
$C_1$ $C_3$ $C_4$ $C_5$        & .5850          & .5826           & .6151          & .5892          & .5826          & .6188          & .5929 \\ \hline
$C_2$ $C_3$ $C_4$ $C_5$         & .5891          & .5822           & .5572          & .5573          & .5875          & .5977          & .5938 \\ \hline
Overall                  & .5825 & .5877  & .5445 & .5791 & .5836 & .5725 & .5780 \\ \hline
\end{tabular}
% \end{table}
\bigskip

\begin{tabular}{|c|c|c|c|c|c|c|c|}
\hline
 \textbf{US-8K}           & $M_S$         & \multicolumn{3}{c|}{Simple TL}                    & \multicolumn{3}{c|}{Neural Adapter TL}           \\ \hline
% \multirow{2}{*}{Labels in $D_S$} &  \multirow{2}{*}{\scriptsize{$D_S$}}    & \multicolumn{3}{c|}{\scriptsize{$D_T$ ($C_1$ $C_2$ $C_3$ $C_4$ $C_5$)}} & \multicolumn{3}{c|}{\scriptsize{$D_T$ ($C_1$ $C_2$ $C_3$ $C_4$ $C_5$)}} \\ \cline{3-8} 
Labels in $D_S$                         & $D_S$           & $D_S$           & New            & All            & $D_S$          & New            & All            \\ \hline

$C_1$ $C_2$ $C_3$ $C_4$          & .6041          & .5857           & .5       

& .5704          & .6023          & .5706          & .5908 \\ \hline
$C_1$ $C_2$ $C_3$ $C_5$          & .5166          & .4523           & .5233          & .5076          & .5047          & .5285          & .5142 \\ \hline
$C_1$ $C_2$ $C_4$ $C_5$          & .6277          & .5568           & .5846          & .5999          & .6145          & .6085          & .6133 \\ \hline
$C_1$ $C_3$ $C_4$ $C_5$         & .6499          & .5865           & .5076          & .5538          & .6192          & .5464          & .5607 \\ \hline
$C_2$ $C_3$ $C_4$ $C_5$          & .6791          & .6756           & .6461          & .6773          & .6675          & .6567          & .6606 \\ \hline
Overall                  & .6154 & .5713  & .5809 & .5818 & .6016 & .5821 & .5871 \\ \hline
\end{tabular}
\vspace{2mm}

\label{tab:us_sed_8k}
\end{table}

\vspace{2mm}

$\bullet$ US-SED: The top table of Table \ref{tab:us_sed_8k} illustrates the results on \textbf{US-SED} dataset. Similar to \textbf{DCASE16}, we can see the neural adapter can effectively bridge the knowledge between the $M_S$ and the $M_T$ models while learning the new sound events. However, by comparing the $D_S$ column in the $M_S$ section to the $D_S$ column in the Simple TL section, we can see the simple TL achieves some improvement over the $M_S$ on the previously learned sound events. Unexpectedly, these results are even slightly better than the neural adapter TL approach (see the $D_S$ column in the Neural Adapter TL section). This result does not match the results obtained for the \textbf{DCASE16} dataset. The reason for this could be the amount of presented noise during the training of the models. As it is mentioned earlier, in the \textbf{US-SED}, some of the used soundscapes in the training data are empty of the sound events of interest. Hence, these files are interpreted as noisy samples by the network. On the other hand, all the sound files used from the \textbf{DCASE16} at least contain one of the desired sound events. To assess this argument, we modify the original UrbanSound8K dataset \cite{salamon2014dataset} to mimic the settings from the \textbf{DCASE16} in creating a more clean dataset\footnote{For our experimental setting, US-8K soundscapes are generated based on the UrbanSound8K dataset which can be found on: \\ \url{https://urbansounddataset.weebly.com/urbansound8k.html}.} and this is the \textbf{US-8K} for our next dataset. In \textbf{US-8K}, each event appears at least once and a maximum of two times in every soundscape file which is the same setting as \textbf{DCASE16} soundscape generation.

\vspace{2mm}

$\bullet$ US-8K: This \textbf{US-8K} dataset has the same five sound events as the \textbf{US-SED}. Similar to \textbf{DCASE16}, each sound event appears in each soundscape file at least once. The evaluation results on this dataset is provided in the bottom table of Table \ref{tab:us_sed_8k}. By looking at the $D_S$ column in the $M_S$ section and the $D_S$ column in the Neural Adapter TL section, it can be seen that similar to the \textbf{DCASE16} dataset, the neural adapter method consistently maintains the performance on the previously learned sound events after learning new sound events. Also transferring the knowledge from the $M_S$ via the neural adapter is more effective compared to the simple TL approach. It is important to ensure that this improvement in the performance is not specific to any target event category, and it is common across different experiments denoted in various rows of the tables. 

\vspace{2mm}

Summarizing the experimental results from the three datasets, it can be seen that simple TL method without the neural adapter has confronted the degradation of the performance with losing the previously learned knowledge from the source model training. There are also up and down in the performance from specific sound events or dataset in the simple TL method. On the other hand, in the case of the neural adapter approach, it is possible to see the inclination in which the learned knowledge is maintained consistently, and the performance sustains in a balanced manner. Therefore, we show performance consistency on the previously learned sound events through neural adapter while obtaining decent detection performance on the newly learned sound events well.

\begin{table}[]
% \small
\centering
\setlength\tabcolsep{1.5pt}

\caption{\textbf{Comparison between individual performances from each model over different settings on DCASE16, US-SED, and US-8K.} \textcircled{A} is the output of the source model with a neural adapter. \textcircled{B} is the output of the target model. \textcircled{C} is a final output of our proposed model (see also Figure \ref{fig:adapter_structure}).}
\vspace{3mm}
\begin{tabular}{|c|c|c|c|}
\hline
                                                        & DCASE16 & US-SED & US-8K \\ \hline
\begin{tabular}[c]{@{}c@{}}\textcircled{A} \end{tabular} & .5321   & .2537  & .3871 \\ \hline
\begin{tabular}[c]{@{}c@{}}\textcircled{B} \end{tabular} & .8017   & .3498  & .5073 \\ \hline
\begin{tabular}[c]{@{}c@{}}\textcircled{C} \end{tabular} & .8451   & .4833  & .5272 \\ \hline
\end{tabular}

\vspace{2mm}

\label{tab:feasibility}
\end{table}

\subsection{The Feasibility of Neural Adapter for Incremental Learning}
\label{sec:adapter_result}

In this section, we study the contribution of each model's outputs separately on the overall performance. In Figure \ref{fig:adapter_structure}, we separate three individual outputs in the neural adapter approach; (1) the optimal output of the source model with a neural adapter (\textcircled{A} in Figure \ref{fig:adapter_structure}), (2) the output of the target model (\textcircled{B} in Figure \ref{fig:adapter_structure}) and (3) the final output of our proposed TL model (\textcircled{C} in Figure \ref{fig:adapter_structure}). For this study, the target domain is used in three cases. The results of the analysis are provided in Table \ref{tab:feasibility}. 

% \vspace{1mm}

It can be seen that when using \textbf{DCASE16} dataset, the trained target model via the neural adapter reaches an optimal point to be able to detect all the \textit{N+1(=4)} sound events without the need to have the source model (see \textcircled{B}\&\textcircled{C} of the DCASE16 column). Therefore, only the target model could be stored and used as a starting point for learning new categories without the need to store the source model and the neural adapter. This result can provide a low footprint continuous learning framework for further model expansion. 

\vspace{2mm}

In contrast, in the case of using the \textbf{US-SED} dataset, the target model and the source model remain complementary to each other for achieving an acceptable final outcome (see \textcircled{B}\&\textcircled{C} of the US-SED column). It appears that the target model cannot maintain the transferred knowledge from the source model while learning the new sound event. Hence, in order to continuously learn new sound events, we always need to keep the whole TL structure presented in Figure \ref{fig:adapter_structure}. This is not feasible for continuous learning in applications that are operated on resource limited platforms, for example, wearable devices. In the case of using the \textbf{US-8K} dataset, it has similar aspect of \textbf{DCASE16}, but it is hard to see a difference as large as \textbf{DCASE16}.

\section{Conclusion and Discussion}

We present an incremental learning algorithm utilizing a TL paradigm for SED application. We use a neural adapter to effectively bridge the gap between the previously learned information in the source model and a target model for learning new sound events. Our extensive analysis shows that utilizing such a mechanism improves the performance of recognizing both known/unknown sound events without forgetting the previously learned knowledge. Thus, our proposed model suits well the scalable and incremental SED applications. 

\vspace{2mm}

This approach can also be used as a low footprint framework for continuous learning in applications that involve less noisy and well annotated data. However, for the more realistic applications, such as acoustic scene classification systems that involve more noisy data, both the target model and the source model might need to remain connected to achieve the desirable performance. Addressing such a challenge remains the focus of our future work.

% \vfill\pagebreak

% References should be produced using the bibtex program from suitable
% BiBTeX files (here: strings, refs, manuals). The IEEEbib.bst bibliography
% style file from IEEE produces unsorted bibliography list.
% -------------------------------------------------------------------------
\bibliographystyle{IEEEbib}
\bibliography{icme2020template}

% \begin{thebibliography}{}
%   \input{icme2020template.bbl}
% \end{thebibliography}

\end{document}